\documentclass[sigconf]{acmart}

\usepackage{booktabs}
\usepackage{algorithm}
\usepackage{algpseudocode}
\usepackage{placeins}

\usepackage{amsmath,amsfonts,bm}

\def\eqref#1{equation~\ref{#1}}

\def\1{\bm{1}}

\DeclareMathAlphabet{\mathsfit}{\encodingdefault}{\sfdefault}{m}{sl}
\SetMathAlphabet{\mathsfit}{bold}{\encodingdefault}{\sfdefault}{bx}{n}

\newcommand{\eq}[1]{Eq.~(\ref{#1})}

\graphicspath{{fig/}}

\setcopyright{acmlicensed}
\acmYear{2026}
\acmConference[DAI 2026]{International Conference on Distributed Artificial Intelligence}{November 29--December 2, 2026}{Hong Kong, China}
\begin{document}

\title{MARA: Flow-Matching-Guided Multi-Agent Resource Allocation for Computational Resource Efficient Learning}

\author{Hanye Zhao}
\email{fineartz@sjtu.edu.cn}
\affiliation{
    \institution{Shanghai Jiao Tong University}
    \city{Shanghai}
    \country{China}
}

\author{Muning Wen}
\email{muningwen@sjtu.edu.cn}
\affiliation{
    \institution{Shanghai Jiao Tong University}
    \city{Shanghai}
    \country{China}
}

\author{Yong Yu}
\email{yyu@apex.sjtu.edu.cn}
\affiliation{
    \institution{Shanghai Jiao Tong University}
    \city{Shanghai}
    \country{China}
}

\author{Weinan Zhang}
\email{wnzhang@sjtu.edu.cn}
\correspondingauthor
\affiliation{
    \institution{Shanghai Jiao Tong University}
    \city{Shanghai}
    \country{China}
}


\begin{abstract}
Allocating limited computation among concurrent learning tasks is difficult when each task must reach a target loss before a deadline but its required training effort is unknown. 
Existing approaches combine online loss prediction with adaptive resource allocation, yet commonly treat computation as continuously divisible throughput. 
We instead study a practical setting in which tasks arrive over time and computation is provided by discrete nodes. 
This setting introduces both uncertain demand and constrained sequential decisions. 
We propose MARA, which predicts future loss trajectories with conditional flow matching and coordinates compute nodes through a cooperative multi-agent autoregressive policy. 
A potential-based progress reward supplies intermediate training feedback while preserving the undiscounted task-completion objective. 
Across in-distribution, reinforcement-learning, and vision workloads, flow matching reduces remaining-resource prediction error relative to weighted least squares. 
At the scheduler's training load, MARA completes 63.46\% of tasks on average, 8.54 percentage points above strong baseline Learning with Adaptive Resource Allocation (LARA), and remains ahead under unseen heavier workloads.
\end{abstract}

\keywords{computational resource efficient learning, multi-agent reinforcement learning, flow matching, learning-curve prediction, resource allocation}

\maketitle

\section{Introduction}
\label{sec:introduction}

\begin{figure*}[t]
  \centering
  \includegraphics[width=0.80\textwidth,trim=0 30 0 30,clip]{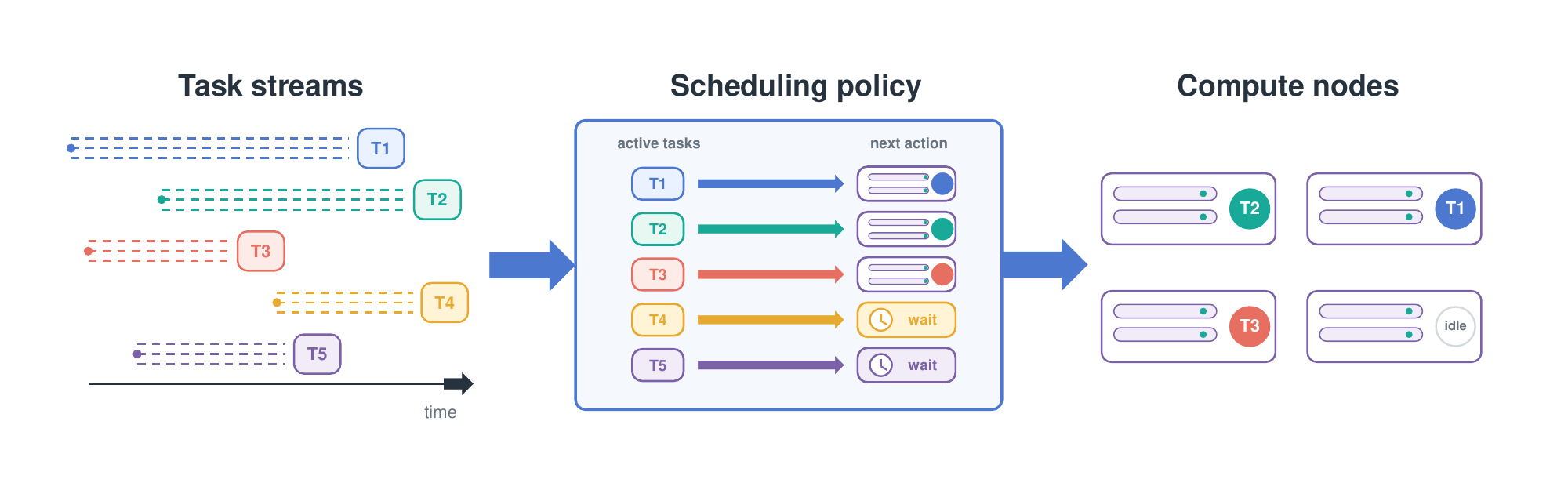}
  \caption{Resource allocation for dynamically arriving task streams.
  The streams have overlapping availability windows and different durations.
  A scheduling policy repeatedly assigns active tasks to discrete compute nodes or leaves them waiting.}
  \Description{Five task streams with staggered arrivals and different durations enter a scheduling policy, which assigns some active tasks to four discrete compute nodes while leaving another task waiting.}
  \label{fig:scenario}
\end{figure*}

Shared learning services often execute many independent training tasks on a limited pool of compute nodes. 
These tasks can differ in architecture, data, optimizer configuration, arrival time, deadline, and target loss. 
A resource allocator must repeatedly decide which tasks should receive computation and which should wait. 
Maximizing utilization or minimizing average completion time is not sufficient in this setting. 
When demand exceeds capacity, spending many batches on a task that cannot finish may prevent several feasible tasks from meeting their deadlines. 
Effective allocation therefore depends on how much training each active task still needs, even though this quantity is unknown when the task arrives. 
Figure~\ref{fig:scenario} illustrates this setting, in which a scheduling policy repeatedly assigns active tasks to discrete compute nodes or leaves them waiting.

This interaction between learning progress and computation is central to Computational Resource Efficient Learning (CoRE-Learning)~\cite{zhou2024core}. 
Within this framework, Learning with Adaptive Resource Allocation (LARA) estimates task demand from observed loss curves and adapts the allocation of divisible data throughput among time-constrained learners~\cite{wang2024lara}. 
We retain its prediction--allocation loop while considering computation supplied by discrete nodes. 
A node serves at most one task in each time step, and a task occupies at most one node. 
Consequently, allocation becomes a constrained task--node matching that must be revised as tasks arrive, finish, or approach their deadlines. 
The matching also determines which tasks produce new loss observations, so prediction and allocation remain coupled throughout an episode.

The discrete setting exposes two related challenges. 
First, demand must be estimated from short, noisy, and heterogeneous loss histories. 
The weighted least-squares (WLS) estimator used by LARA is efficient, but its fixed negative-power form cannot represent every plateau or change in convergence rate. 
A single extrapolated value also hides the different futures that may agree with the same observed prefix. 
Second, several node decisions must be coordinated at every time step. 
Assigning a task to one node removes it from the choices available to the remaining nodes, while a locally attractive assignment may consume capacity needed to complete a better set of tasks.
Learning such behavior from task completions alone is difficult because many allocation decisions precede each positive reward.

We address these challenges with Multi-Agent Resource Allocation (\emph{MARA}), a resource-allocation method that combines conditional flow matching with cooperative multi-agent reinforcement learning. 
Given an observed prefix, the predictor generates a distribution over future loss trajectories~\cite{lipman2023flow}. 
Threshold crossings in these trajectories yield online estimates of remaining training demand without tying the learned representation to one target loss. 
The resulting estimates enter a Multi-Agent Transformer (MAT), which assigns tasks to nodes autoregressively~\cite{wen2022mat}. 
Allocation masks enforce valid matchings, and a log-progress potential supplies intermediate feedback while preserving the original undiscounted objective of completing as many tasks as possible.

Our evaluation examines both parts of this design. 
We first compare FM and WLS as remaining-demand predictors, and then evaluate end-to-end allocation under in-distribution tasks, held-out task families, and increasing arrival loads. 
Improved trajectory forecasts translate into better allocation, while targeted stress tests and operational diagnostics clarify when coordinated decisions are most useful.

Our contributions are as follows:

\begin{itemize}
    \item To our knowledge, MARA is the first multi-agent formulation of CoRE-Learning problem with dynamic arrivals, deadlines, and initially unknown training demand.
    \item We introduce a conditional flow-matching predictor that generates future loss trajectories and decodes threshold crossings into online remaining-resource estimates.
    \item We develop an autoregressive MAT allocator with feasibility masks and an objective-preserving log-progress potential for sparse completion feedback.
    \item MARA achieves the highest completion rate among the evaluated realizable methods, exceeding LARA by 8.54 points at the training load and by 8.43--8.84 points under heavier loads. 
\end{itemize}

\section{Related Work}
\label{sec:related}

\paragraph{Computational resource efficient learning}
CoRE-Learning formalizes learnability when multiple learners share limited, time-varying computational throughput~\cite{zhou2024core}.
LARA turns this principle into an online allocator that uses WLS loss-curve extrapolation to balance information collection and resource assignment~\cite{wang2024lara}.
These studies make learning progress part of the allocation state, rather than assuming that a training task has known resource demand.
We retain this view while considering discrete compute nodes and coupled task assignments.

\paragraph{Learning-curve prediction}
Learning curves have been used to stop weak configurations early and direct resources toward promising ones.
Domhan et al. extrapolate partial curves with a mixture of parametric models~\cite{domhan2015speeding}, while Klein et al. transfer curve information across configurations and datasets with a Bayesian neural model~\cite{klein2017learning}.
Hyperband instead frames early stopping as adaptive resource allocation~\cite{li2018hyperband}.
LARA's WLS estimator is attractive online because it is inexpensive, but a fixed parametric curve can be restrictive.
Flow matching learns a continuous transport from a simple distribution to a target distribution~\cite{lipman2023flow}, while rectified flow emphasizes paths that admit efficient numerical integration~\cite{liu2023flow}.
These properties motivate forecasting a distribution over future loss trajectories before decoding the remaining demand.

\paragraph{Learning-aware scheduling}
Several cluster schedulers use training progress to improve resource decisions.
SLAQ prioritizes jobs by predicted quality improvement~\cite{zhang2017slaq}, Optimus adjusts distributed workers from online progress estimates~\cite{peng2018optimus}, and Pollux co-adapts job-level training choices and cluster-level allocation to optimize goodput~\cite{qiao2021pollux}.
Together, they show that learning dynamics can be useful scheduling signals when demand is not known in advance.
Our setting instead assigns indivisible nodes to single-node tasks with individual loss targets and deadlines.

\paragraph{Cooperative MARL}
Cooperative MARL provides several ways to represent coupled decisions.
QMIX factorizes a joint action value under a monotonicity constraint~\cite{rashid2018qmix}, and MAPPO shows that a carefully implemented on-policy method can be competitive across cooperative benchmarks~\cite{yu2022mappo}.
MARL has also coordinated job ordering and GPU placement in deep-learning clusters~\cite{xing2023dual}, but that setting optimizes job cost and completion time rather than loss-target completion under unknown demand.
MAT models the joint policy autoregressively and connects sequence modeling with multi-agent advantage decomposition~\cite{wen2022mat}.
We use this factorization for ordered node assignments and train it with PPO and GAE~\cite{schulman2017ppo,schulman2016gae}.
Potential-based shaping supplies intermediate feedback while preserving the underlying objective under the stated boundary conditions~\cite{ng1999policy}.

\section{Problem Formulation}
\label{sec:problem}

We first define task success and feasible allocation, and then express their interaction as a cooperative multi-agent decision problem.
Consider a set of homogeneous compute nodes $\mathcal{N}=\{1,\ldots,M\}$ and discrete decision times $t=0,1,\ldots$.
Learning task $k$ arrives at time $b_k$, has deadline $d_k$, and specifies a target loss $\epsilon_k$.
Its training recipe fixes the data, model, optimizer, batch size, and learning rate.
Let $s_k(t)$ be the number of optimization batches processed by time $t$, and let $L_k(s)$ be its cumulative-average training loss after $s$ batches.
Because $L_k$ is unknown before execution, success can only be determined from observations collected online.
Task $k$ succeeds if its loss reaches the target threshold before deadline, i.e. $L_k(s_k(t))\leq\epsilon_k$ for some $t\leq d_k$.

Given this success criterion, the allocator must choose which active tasks receive the limited nodes.
At time $t$, the active set $\mathcal{A}_t$ contains arrived tasks that have neither succeeded nor expired.
Node $i$ selects
\begin{equation}
  a_{i,t}\in\mathcal{A}_t\cup\{\mathrm{idle}\}.
\end{equation}

\begin{figure*}[t]
  \centering
  \includegraphics[width=\textwidth,trim=0 20 0 20,clip]{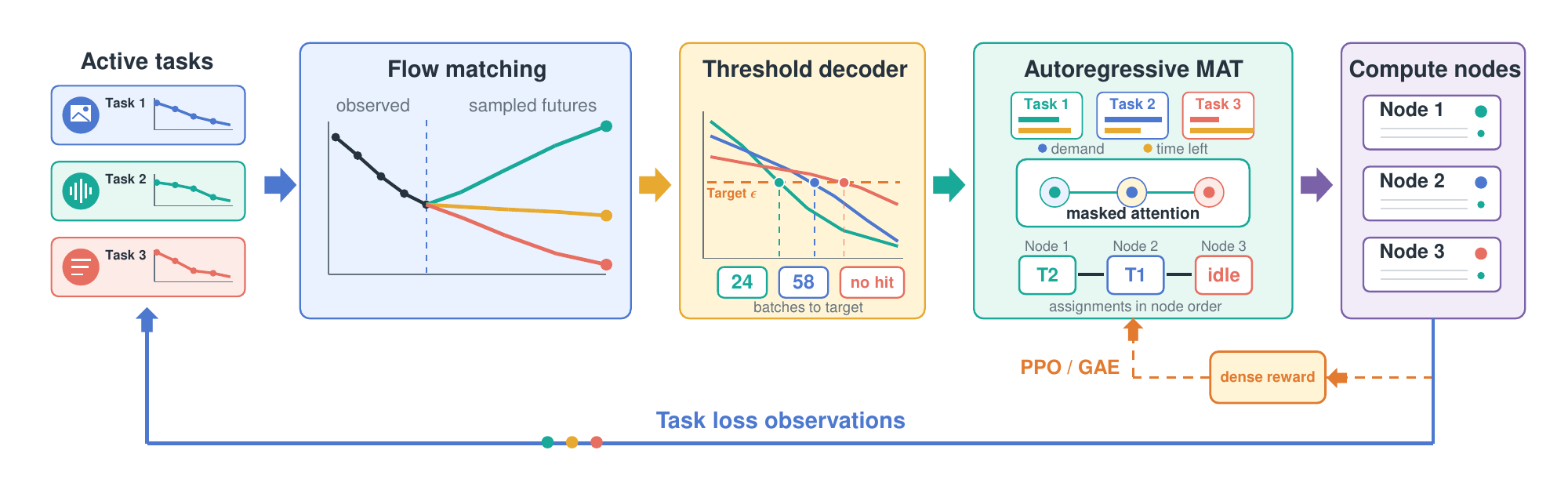}
  \caption{Overview of MARA.
  Observed loss histories condition the flow-matching predictor, while decoded resource estimates and task state enter the autoregressive MAT allocator.
  Simulator transitions provide new loss observations and shaped training rewards.}
  \Description{A left-to-right pipeline begins with three active tasks and their observed loss curves.
  A flow-matching model samples future trajectories, and a threshold decoder converts their crossing points into resource-demand tokens.
  An autoregressive MAT policy uses these tokens to order assignments to three compute nodes.
  A blue loop returns new loss observations to the tasks, while an orange dashed loop sends training rewards through PPO and GAE to MAT.}
  \label{fig:overview}
\end{figure*}

The joint action must satisfy
\begin{equation}
  \sum_{i\in\mathcal{N}}\mathbb{I}[a_{i,t}=k]\leq 1,
  \qquad \forall k\in\mathcal{A}_t,
  \label{eq:matching}
\end{equation}
so each node serves at most one task and each task uses at most one node.
For each timestep, a selected task executes one optimization batch.
Tasks may be preempted and resumed between decisions, and the present model assumes no switching or migration cost.
The allocation objective is to maximize the number of successful tasks:
\begin{equation}
  \max_{\pi}\;J(\pi)=
  \mathbb{E}_{\pi}\!\left[\sum_k
  \mathbb{I}\!\left(\exists t\leq d_k:
  L_k(s_k(t))\leq\epsilon_k\right)\right].
  \label{eq:objective}
\end{equation}

The objective in \eq{eq:objective} couples node decisions through both shared capacity and future task progress.
We therefore cast the problem as a cooperative partially observed Markov game.
Each node is an agent, and all agents share the completion reward.
At a decision time, the policy observes the active task slots, their loss histories, predicted remaining batches, and remaining deadlines, but not their future loss curves.
The agents act in a fixed node-index order to construct a matching.
Although nodes are homogeneous, earlier actions restrict later choices through \eq{eq:matching}, and the autoregressive factorization represents this dependence without enumerating the combinatorial set of feasible matchings.

\section{MARA}
\label{sec:method}

MARA closes the loop between loss prediction and resource allocation.
Figure~\ref{fig:overview} illustrates the overall MARA framework.
When a task has accumulated sufficient history, a conditional flow model generates future loss segments, and a crossing decoder estimates its remaining batches.
These estimates, together with deadlines and history progress, form task tokens for a MAT policy.
MAT assigns tasks to nodes autoregressively, and the simulator executes one batch for each selected task.
The resulting losses update the next prediction and decision, while predictor parameters remain fixed during scheduler training.

\subsection{Future-Loss Prediction with Flow Matching}
\label{sec:flow}

We first build history representations for tasks.
For task $k$, the observed prefix records the cumulative processed samples and cumulative-average losses.
To normalize different lengths and loss scales, we encode each point using the log processed-sample count, log loss relative to the current loss, and their first differences.
A GRU then maps this variable-length sequence to a history representation.
We concatenate four observable static features---batch size, learning rate, current loss, and forecast horizon---with that representation.
We exclude the target threshold $\epsilon_k$ so that the predicted loss curve remains threshold-independent.

The predictor also needs a fixed-length representation of the variable future horizon.
Therefore, we construct a target vector $\mathbf{y}_k\in\mathbb{R}^{H}$ with $H=32$.
Let $S_0$ be the processed samples at the end of the prefix, let $h$ be the queried future horizon in samples, and define equal-width boundaries
\begin{equation}
  s_j=S_0+\frac{jh}{H},\qquad j=0,\ldots,H.
  \label{eq:bin-boundaries}
\end{equation}
Because $L(s)$ is a cumulative-average loss, $C(s)=sL(s)$ is the cumulative loss sum.
We linearly interpolate $C$ at every $s_j$ and form the exact interval average
\begin{equation}
  \bar{\ell}_j=\frac{C(s_j)-C(s_{j-1})}{s_j-s_{j-1}},\qquad
  y_j=\log\frac{\bar{\ell}_j+\delta}{L(S_0)+\delta},
  \label{eq:interval-target}
\end{equation}
where $\delta$ is a small numerical constant.
The inverse transform gives $\hat{\bar{\ell}}_j$ and reconstructs
\begin{equation}
  \hat C(s_j)=C(S_0)+\sum_{r=1}^{j}(s_r-s_{r-1})\hat{\bar{\ell}}_r,
  \qquad \hat L(s_j)=\frac{\hat C(s_j)}{s_j}.
  \label{eq:curve-reconstruction}
\end{equation}
This construction preserves the observed cumulative sum at $S_0$ and gives all 32 coordinates the same sample-space interpretation.
Further, telative log targets can reduce scale variation across different task families.

This trajectory target is preferable to directly regressing remaining batches for two reasons.
First, remaining batches depend on the query threshold and are right-censored when a curve does not cross within the generation cap, whereas a future segment supplies dense supervision even in these cases.
Second, a single generated segment can be decoded for different thresholds and horizons without retraining the predictor.
Consequently, we can train this predictor separetely from the scheduler policy.

Given this target representation, conditional flow matching models the distribution of plausible future segments.
Let $\mathbf{z}_1$ be the standardized target and draw $\mathbf{z}_0\sim\mathcal{N}(\mathbf{0},\mathbf{I})$.
For flow time $u\sim\mathcal{U}[0,1]$, we use the linear conditional path
\begin{equation}
  \mathbf{z}_u=(1-u)\mathbf{z}_0+u\mathbf{z}_1.
\end{equation}
A velocity network $v_\theta(\mathbf{z}_u,u,\mathbf{c}_k)$ receives the current state, a sinusoidal time embedding, and encoded context $\mathbf{c}_k$.
It is trained with
\begin{equation}
  \mathcal{L}_{\mathrm{FM}}(\theta)=
  \mathbb{E}\left[
  \left\|v_\theta(\mathbf{z}_u,u,\mathbf{c}_k)
  -(\mathbf{z}_1-\mathbf{z}_0)\right\|_2^2
  \right].
  \label{eq:fm}
\end{equation}
For low-latency scheduling, we use a lightweight GRU history encoder and an MLP velocity field.
At inference, we integrate $d\mathbf{z}/du=v_\theta(\mathbf{z},u,\mathbf{c}_k)$ with a small number of Euler steps.
Fixed Sobol Gaussian points make repeated predictions deterministic for the same prefix.
The detailed architecture and sampling settings are reported in Appendix~\ref{app:implementation}.

Finally, the allocator requires one remaining-demand estimate for each task.
To obtain it, we reconstruct the future cumulative-average loss for each sampled trajectory and find its first crossing of $\epsilon_k$.
Linear interpolation within the crossing interval yields the remaining processed samples, which we convert to optimization batches.
If a trajectory never crosses, its estimate is capped at the generation horizon.
The predictor samples $N$ initial noise vectors, decodes each trajectory into a remaining-batch estimate, and returns their median.
This aggregation provides a robust scalar estimate while preserving multimodal future behavior.
Before four history points are available, the predictor conservatively returns the remaining generation cap.

\subsection{Autoregressive Multi-Agent Allocation}
\label{sec:mat}

The decoded demand estimates connect future-loss prediction to task--node matching.
Because two nodes cannot select the same active task, we formulate matching as a multi-agent sequential decision problem and solve it with the Multi-Agent Transformer (MAT).
Each compute node acts as a MAT agent, while the active set occupies at most $K$ task slots with zero padding.
Every agent receives the same global observation, which contains normalized node and active-task counts, predicted remaining batches, remaining deadlines, and the fraction of the four-point prediction warm-up completed for each slot.
The warm-up fraction identifies tasks whose demand estimates may still be inaccurate.
A Transformer encoder models competition among these task features and produces a centralized value estimate.

The decoder constructs $p_\psi(\mathbf{a}_t\mid o_t)$ in fixed node-index order:
\begin{equation}
  p_\psi(\mathbf{a}_t\mid o_t)
  =\prod_{i=1}^{M}p_\psi(a_{i,t}\mid o_t,a_{1,t},\ldots,a_{i-1,t}).
  \label{eq:autoregressive}
\end{equation}
At step $i$, a mask removes padded slots and all tasks selected by previous nodes.
These masks guarantee \eq{eq:matching} without repair after sampling.
Autoregressive decoding captures dependencies among node choices without enumerating all possible matchings.
The multi-agent advantage decomposition theorem applies to any agent ordering and supports this sequential factorization~\cite{kuba2022trust,wen2022mat}.

Having defined a valid joint policy, we optimize it with PPO~\cite{schulman2017ppo} and use GAE~\cite{schulman2016gae} to estimate its advantages.
Let $r_t(\psi)=\pi_\psi(\mathbf{a}_t\mid o_t)/
\pi_{\psi_{\mathrm{old}}}(\mathbf{a}_t\mid o_t)$ and let $\hat A_t$ be the GAE advantage.
The clipped policy objective is
\begin{equation}
  \mathcal{L}_{\mathrm{clip}}(\psi)=
  \mathbb{E}_t\!\left[\min\!\left(
  r_t(\psi)\hat A_t,
  \operatorname{clip}(r_t(\psi),1-\varepsilon_{\mathrm{PPO}},
  1+\varepsilon_{\mathrm{PPO}})\hat A_t\right)\right].
  \label{eq:ppo-clip}
\end{equation}
For return target $\hat R_t$, the value loss and entropy are
\begin{equation}
  \mathcal{L}_V=\tfrac12\mathbb{E}_t[(V_\phi(o_t)-\hat R_t)^2],
  \qquad
  \mathcal{H}=\mathbb{E}_t[\mathcal{H}(\pi_\psi(\cdot\mid o_t))].
\end{equation}
We minimize
\begin{equation}
  \mathcal{L}_{\mathrm{MAT}}=-\mathcal{L}_{\mathrm{clip}}
  +c_v\mathcal{L}_V-c_e\mathcal{H},
  \label{eq:ppo-total}
\end{equation}
with the coefficients and GAE settings reported in Appendix~\ref{app:implementation}.

\subsection{Objective-Preserving Progress Shaping}
\label{sec:shaping}

Although PPO can optimize the joint policy above, the original objective provides a unit reward only when a task succeeds.
We therefore use potential-based reward shaping to expose intermediate progress.
For an active task with initial loss $L_{k,0}$ and current loss $L_{k,t}$, define
\begin{equation}
  q_k(t)=\operatorname{clip}\!\left(
  \frac{\log\!\left(L_{k,0}/\max(L_{k,t},\epsilon_k)\right)}
       {\log\!\left(L_{k,0}/\epsilon_k\right)},0,1\right),
  \label{eq:progress}
\end{equation}
with $q_k(t)=1$ if the initial loss already meets the threshold.
The state potential is $\Phi_t=\sum_{k\in\mathcal{A}_t}q_k(t)$.
Our training reward is
\begin{equation}
  r_t=r_t^{\mathrm{success}}+\beta(\Phi_{t+1}-\Phi_t).
  \label{eq:shaping}
\end{equation}
Successful and expired tasks are removed before $\Phi_{t+1}$ is evaluated, and new tasks begin with zero progress.
At both episode boundaries the active set is empty, hence $\Phi_0=\Phi_T=0$.
With discount factor $\gamma=1$,
\begin{equation}
  \sum_{t=0}^{T-1}r_t
  =\sum_{t=0}^{T-1}r_t^{\mathrm{success}}
   +\beta(\Phi_T-\Phi_0)
  =\sum_k\mathbb{I}[k\text{ succeeds}].
\end{equation}
The dense signal therefore changes credit assignment without changing the undiscounted episode objective.
These rewards are used only for policy training, while inference requires neither rewards nor the true future curve.

\subsection{Training and Workload Randomization}

FM and MAT rely on different supervision, so we train them in two stages.
For FM, we collect complete curves from eight in-distribution (ID) model--dataset configurations and run each task to a finite generation cap, including after it first reaches the target.
We randomize initialization, data order, batch size, and learning rate to expose the predictor to varied convergence behavior.
An eligible prefix of at least four points and a horizon contained in the stored suffix define one training example through Eqs.~(\ref{eq:bin-boundaries})--(\ref{eq:interval-target}).
Dataset sizes and randomization ranges are given in Appendix~\ref{app:implementation}.
In the second stage, the selected predictor is frozen while MAT interacts with a simulator of task arrivals, training, and node assignment.
Scheduler workloads randomize arrival patterns, task families, task seeds, target thresholds, and per-family scheduling budgets.
This separation reflects the supervision available to each module.
Full future curves directly supervise the predictor, whereas scheduler rollouts reveal only the observations selected by the current policy.
Joint updates would make the policy input non-stationary, while freezing FM makes the WLS--FM comparison a controlled change of predictor.
Algorithm~\ref{alg:mara} summarizes the complete procedure.

\begin{algorithm}[t]
  \caption{Staged training and online allocation in MARA}
  \label{alg:mara}
  \small
  \begin{algorithmic}[1]
    \Require ID task configurations $\mathcal{C}$, simulator $\mathcal{E}$, node count $M$
    \Statex \textbf{Stage I: train the trajectory predictor}
    \For{$c\in\mathcal{C}$ and randomized task seeds/recipes}
      \State Run $c$ to the finite generation cap and store complete $L(s)$
    \EndFor
    \Repeat
      \State Sample a training curve, prefix $S_0$, and valid horizon $h$
      \State Build context $\mathbf{c}$ and 32-bin target $\mathbf{y}$ by \eq{eq:interval-target}
      \State Update $\theta$ using $\mathcal{L}_{\mathrm{FM}}$ in \eq{eq:fm}
    \Until{validation loss stops improving}
    \State Freeze the selected predictor $v_\theta$
    \Statex \textbf{Stage II: train the allocator}
    \Repeat
      \State Reset $\mathcal{E}$ with a randomized task stream
      \While{the episode is active}
        \ForAll{active tasks $k$}
          \State Set $\hat b_k$ from FM if ready, and use the cap otherwise
        \EndFor
        \State Sample a masked autoregressive task--node matching $\mathbf{a}_t$
        \State Execute selected batches and observe shaped reward $r_t$
      \EndWhile
      \State Compute GAE and update actor/critic with \eq{eq:ppo-total}
    \Until{the MAT training budget is exhausted}
    \Statex \textbf{Online use:} repeat the prediction, masked matching, and execution steps without parameter updates.
  \end{algorithmic}
\end{algorithm}

\section{Experiments}
\label{sec:experiments}

Our experiments separate predictor quality from end-to-end allocation while preserving the interaction between them.
All allocators process matched workloads, and every learned policy is evaluated on frozen task curves that remain hidden beyond the observations produced online.
We first define the tasks, workloads, methods, and metrics, and then study prediction, allocation, generalization, operational behavior, and controlled ablations.
The evaluation addresses four questions.

\begin{samepage}
\begin{enumerate}
  \item[\textbf{RQ1}] Does FM predict remaining training demand more accurately than WLS?
  \item[\textbf{RQ2}] At the load used to train MAT, does MARA complete more tasks than the allocation baselines?
  \item[\textbf{RQ3}] How well does MARA generalize to held-out task families and unseen heavier workloads?
  \item[\textbf{RQ4}] How do prediction quality, allocation policy, and operational behavior contribute to the end-to-end result?
\end{enumerate}
\end{samepage}

\subsection{Experimental Setup}

\paragraph{Tasks and parameterization}
We inherit the task models and datasets from the pure and mixed bundles in LARA~\cite{wang2024lara}.
The ID set contains four CIFAR-10 classifiers~\cite{krizhevsky2009cifar}, two Transformer models on the YESNO speech corpus~\cite{openslrYesno}, and two attention-LSTM classifiers on IMDB~\cite{maas2011imdb}.
RL OOD introduces a DAgger-style CNN trained on expert image--action pairs from Atari Montezuma's Revenge through the Arcade Learning Environment~\cite{bellemare2013ale}, while ViT OOD introduces a SimpleViT classifier on CIFAR-10.
Neither OOD family appears in FM or MAT training.
Together, these settings cover image, text, speech, and RL tasks with widely used models.
Although the model--dataset combinations follow LARA, their temporal and success parameters are recalibrated for discrete service.
Available time $B$ is an integer budget under which a continuously served task can execute at most $B$ optimization batches before its deadline.
Detailed task configurations and sampling ranges are given in Appendix~\ref{app:implementation}.

\paragraph{Workloads and simulator}
Using these task definitions, we construct streams with both arrival and task-level variation.
ID arrivals are sampled uniformly from the eight ID configurations.
The RL OOD and ViT OOD task sets independently replace 20\% of these arrivals with \texttt{RLExpertConfig} and \texttt{ViT\_Cifar10\_1}, respectively.
They are therefore mixed task sets rather than OOD-only episodes.
Task targets, service budgets, initialization seeds, and data orders are randomized when workloads and complete curves are generated.
Each evaluation uses seven homogeneous nodes, and a selected task executes one optimization batch.
MAT is trained only with arrival probability $p=0.04$.
We use this training load for the primary comparison and test $p\in\{0.05,0.06,0.07\}$ only to measure zero-shot load generalization.
For every load and task set, all systems process the same 12 episodes for evaluation.
Across the five evaluation roots, these episodes contain about $19.70$, $23.82$, $27.75$, and $31.75$ tasks on average at $p=0.04,0.05,0.06,0.07$, respectively, after giving equal weight to the three task sets. 
The predictor and MAT configurations, training budgets, checkpoint rule, and simulator limits are reported in Appendix~\ref{app:implementation}.

\paragraph{Evaluation design}
The evaluation setup separates task-family generalization from load generalization.
Fixing $p=0.04$ for the primary comparison keeps the evaluation aligned with the distribution used to optimize MAT.
Averaging all four loads into that comparison would mix performance on the training distribution with extrapolation to distributions that the policy never observed.
The ID, RL OOD, and ViT OOD conditions measure task-family shifts under the same arrival process, whereas the heavier-load evaluation changes arrival frequency while keeping the trained checkpoint fixed.
Pre-generating workloads further separates policy quality from workload sampling because each evaluation root fixes arrivals, task parameters, and the complete loss curves used by the simulator.
Every policy nevertheless observes only the prefix produced by its own actions.

\paragraph{Methods}
We compare MARA with several baselines and controlled variants.
\textbf{MARA} denotes FM+MAT with the dense reward in \eq{eq:shaping}.
\textbf{MARA-WLS} replaces FM with WLS while retaining the same MAT allocator and dense reward.
\textbf{LARA} adapts the original throughput allocator to discrete nodes and retains its WLS predictor and adaptive binary-tree search.
When tasks lack sufficient prediction history, at most one node is reserved to collect observations.
\textbf{LARA-FM} changes only LARA's predictor from WLS to FM.
\textbf{FIFO} and \textbf{EDF} are work-conserving first-in-first-out and earliest-deadline-first policies that serve as naive baselines.
\textbf{Oracle} is an independent-feasibility upper bound that counts a task if its fixed loss curve can reach the threshold in isolation while ignoring node competition, queueing, and admission.
It is not a realizable policy and is used only to characterize workload difficulty.
All comparisons share workload seeds, task order, curves, constraints, and available online information.

\paragraph{Metrics}
For prediction, capped remaining-batch MAE measures numerical demand error and clips non-crossing predictions at the forecast horizon.
Budget-success accuracy is the fraction of prefixes for which the predictor correctly determines whether the task can finish within its scheduling budget.
Inference latency is the wall-clock prediction time for one prefix.
For allocation, the success rate is the fraction of tasks that reach their loss targets before their deadlines.
We compute this fraction in each episode and average the 12 episode values equally to obtain a task-set mean.
The overall average then gives equal weight to the three task-set means.
The same two-stage calculation is applied independently at every load, which prevents task sets with more generated tasks from dominating the aggregate.
For MARA and MARA-WLS, the reported mean and sample standard deviation use all 15 training-seed--evaluation-root combinations.
LARA-FM, LARA, FIFO, EDF, and Oracle do not depend on a MAT training seed, so their statistics use the five evaluation roots.

\subsection{Resource Prediction}
\label{sec:prediction-results}

We first test whether FM provides the allocator with more reliable demand estimates than WLS.
Capped MAE measures the numerical error in remaining batches, while budget-success accuracy evaluates the resulting binary feasibility decision.

\begin{table}[t]
  \caption{Remaining-resource prediction.
  MAE is in optimization batches.
  Acc. is scheduling-budget success accuracy.}
  \label{tab:prediction}
  \centering
  \begin{tabular}{@{}llccc@{}}
    \toprule
    Task set & Model & MAE $\downarrow$ & Acc. (\%) $\uparrow$ & ms/prefix $\downarrow$ \\
    \midrule
    ID & WLS & 92.37 & 82.67 & 0.054 \\
       & FM  & \textbf{44.23} & \textbf{90.33} & 0.252 \\
    RL OOD & WLS & 242.43 & 51.25 & 0.054 \\
           & FM  & \textbf{141.62} & \textbf{70.00} & 0.262 \\
    ViT OOD & WLS & 82.09 & 84.17 & 0.053 \\
            & FM  & \textbf{79.37} & \textbf{85.00} & 0.255 \\
    \bottomrule
  \end{tabular}
\end{table}

\begin{figure*}[t]
  \centering
  \includegraphics[width=\textwidth]{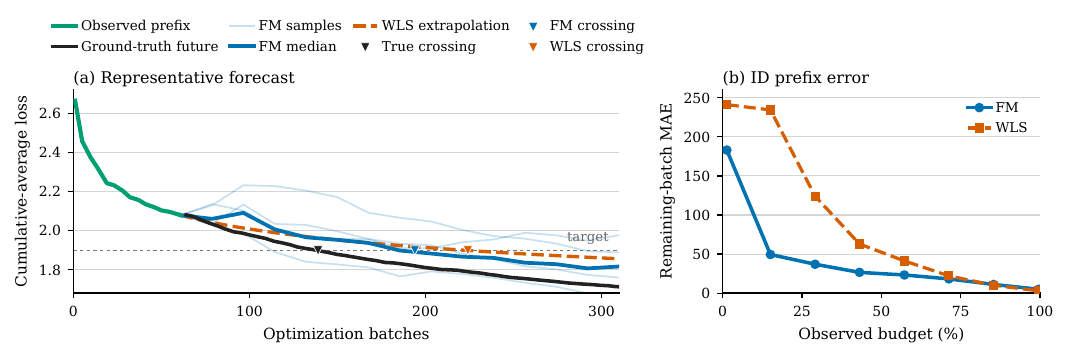}
  \caption{Prediction behavior on ID tasks as observations accumulate. (a) A representative curve, selected by the median curve-level WLS-minus-FM MAE improvement, at its fourth query prefix. (b) Capped remaining-batch MAE at eight deterministic query ranks.
  The horizontal coordinate is the median observed fraction of the scheduling budget.}
  \Description{Two panels compare FM and WLS prediction on ID tasks.
  The first shows an observed loss prefix, the ground-truth suffix, sampled FM futures, the FM median, the WLS extrapolation, and three threshold crossings.
  The second plots remaining-batch error as the observed fraction of the task budget grows.}
  \label{fig:prediction}
\end{figure*}

The two predictive metrics show the same overall pattern in Table~\ref{tab:prediction}.
FM reduces capped MAE by 52.1\% on ID tasks and 41.6\% on RL OOD tasks, while success classification improves by 7.66 and 18.75 percentage points, respectively.
The gain on ViT is smaller but remains positive, with a 3.3\% reduction in MAE and a 0.83-point increase in success accuracy.
FM takes roughly 0.25 ms per prefix, compared with 0.05 ms for WLS in this batched benchmark on an NVIDIA H20 GPU.
Although FM is slower, the additional prediction time is negligible compared with task training time.

Figure~\ref{fig:prediction} shows when the predictors diverge as a task reveals more of its curve by pairing a representative ID forecast with remaining-batch errors across successive query points.
Panel (a) makes the threshold-crossing error concrete.
The true curve reaches the target at batch 139.
From the prefix observed through batch 61, FM predicts batch 194, whereas WLS predicts batch 224.07.
Both estimates are conservative, but FM reduces the absolute crossing error from 85.07 to 55 batches.
Panel (b) shows that this is not only a late-curve effect.
FM has substantially lower MAE through the first six query ranks, when at most about 71\% of the budget has been observed and scheduling decisions still affect resource use.
Once nearly the entire budget is visible, both errors become small, and WLS is slightly lower at the final rank.
FM therefore provides its main benefit during the early and middle prefixes, when long-range uncertainty remains consequential for allocation.

\subsection{Main Allocation Results}
\label{sec:scheduling-results}

We next evaluate whether the prediction and coordination components improve end-to-end completion at the load used to train MAT.
Table~\ref{tab:main} reports the three task sets separately and then averages them with equal weight.

\begin{table*}[t]
  \caption{Main allocation results at the MAT training load $p=0.04$.
  Cells report mean $\pm$ sample standard deviation of success rate (\%).
  Average gives equal weight to the three task sets.}
  \label{tab:main}
  \centering
  \begin{tabular*}{\textwidth}{@{\extracolsep{\fill}}lccccccc@{}}
    \toprule
    Task set & MARA & MARA-WLS & LARA & LARA-FM & FIFO & EDF & Oracle \\
    \midrule
    ID & \textbf{61.90 $\pm$ 4.09} & 60.54 $\pm$ 3.10 & 53.60 $\pm$ 4.02 & 59.05 $\pm$ 4.24 & 51.54 $\pm$ 4.09 & 51.57 $\pm$ 3.64 & 68.06 $\pm$ 2.27 \\
    RL OOD & \textbf{65.31 $\pm$ 3.18} & 63.61 $\pm$ 3.04 & 56.35 $\pm$ 3.08 & 62.41 $\pm$ 3.80 & 54.75 $\pm$ 2.90 & 54.86 $\pm$ 2.23 & 71.69 $\pm$ 1.70 \\
    ViT OOD & \textbf{63.18 $\pm$ 3.92} & 62.07 $\pm$ 3.01 & 54.83 $\pm$ 4.26 & 60.45 $\pm$ 4.42 & 52.90 $\pm$ 4.25 & 53.83 $\pm$ 3.61 & 71.12 $\pm$ 2.54 \\
    \midrule
    Average & \textbf{63.46 $\pm$ 3.61} & 62.07 $\pm$ 2.91 & 54.93 $\pm$ 3.69 & 60.64 $\pm$ 4.09 & 53.06 $\pm$ 3.61 & 53.42 $\pm$ 3.10 & 70.29 $\pm$ 2.11 \\
    \bottomrule
  \end{tabular*}
\end{table*}

MARA obtains the highest mean among realizable methods on every task set at the load on which MAT was trained (Table~\ref{tab:main}).
Its average success rate is 63.46\%, compared with 54.93\% for LARA, for a gain of 8.54 percentage points.
The gains over the strongest main baseline are 8.30 points on ID, 8.96 points on RL OOD, and 8.35 points on ViT OOD.
MARA also exceeds the MARA-WLS ablation by 1.39 points on average.

Oracle is not attainable under the shared-node constraint, but its gap to MARA separates failures caused by intrinsic task infeasibility from those caused by allocation and congestion.
The remaining mean gaps are 6.15, 6.38, and 7.94 points on ID, RL OOD, and ViT OOD, respectively.
The held-out task sets also have different Oracle values, so their absolute success rates should not be read as a direct ranking of task-set difficulty.
Comparisons within the same task set are more informative because they hold task composition and intrinsic feasibility fixed while changing only the allocation method.

\subsection{Infeasible-Task Stress Test}
\label{sec:trap-analysis}

Average workloads may hide a specific scheduling failure in which tasks that cannot finish continue to receive service while feasible tasks wait.
We isolate this behavior with a deliberately constructed stress test.

\begin{table}[t]
  \caption{Targeted infeasible-task stress test.
  Cells report mean $\pm$ sample standard deviation of success rate (\%).}
  \label{tab:trap}
  \centering
  \begin{tabular}{@{}lc@{}}
    \toprule
    Method & Success rate (\%) \\
    \midrule
    MARA & \textbf{$65.00\pm0.00$} \\
    MARA-WLS & $57.83\pm8.23$ \\
    LARA & $11.00\pm8.40$ \\
    LARA-FM & $45.50\pm17.45$ \\
    FIFO & $0.00\pm0.00$ \\
    EDF & $0.00\pm0.00$ \\
    Oracle & $65.00\pm0.00$ \\
    \bottomrule
  \end{tabular}
\end{table}

This stress test targets wasted service on tasks that cannot meet their objectives.
Seven tasks arrive at time 0 with loss curves that never cross their thresholds within their service budgets.
At time 10, thirteen short tasks arrive whose thresholds can be reached in 22--35 batches.
The early tasks initially occupy all seven nodes and have earlier deadlines, so policies that cannot properly estimate the required resources will continue serving them while the feasible tasks wait.
As a result, Table~\ref{tab:trap} shows a sharp separation.
MARA completes all thirteen feasible tasks and matches the 65\% independent-feasibility bound with zero observed variation.
MARA-WLS remains effective but is less stable, while replacing MAT with the fixed LARA allocator lowers the rate even when it uses FM.
FIFO and EDF complete no task because they spend the available service window on the early infeasible set.
The result isolates MARA's ability to combine demand estimates with coordinated decisions about when to stop investing in an unpromising task.

\subsection{Generalization to Held-Out Task Sets}
\label{sec:task-generalization}

\paragraph{RL OOD}
At the fixed training load, FM transfers differently to RL and ViT loss curves.
RL replaces only one fifth of arrivals, so each result mixes held-out and ID behavior.
Even with this dilution, the predictor result provides a clear mechanism because FM lowers capped remaining-batch MAE from 242.43 to 141.62 and raises budget-success accuracy from 51.25\% to 70.00\%.
The RL loss trajectories often continue improving after prefixes for which the local WLS fit is pessimistic.
Forecasting a nonlinear future segment recovers more of these threshold crossings.
End to end, MARA reaches 65.31\%, which is 8.96 points above LARA and 1.70 points above MARA-WLS.
Its 6.38-point mean Oracle gap indicates that most independently feasible tasks are completed despite the task-family shift.

\paragraph{ViT OOD}
The ViT result differs.
FM improves capped MAE by only 3.3\% and budget-success accuracy by 0.83 points over WLS.
The held-out SimpleViT curves show a slower cumulative-loss decline, making early prefixes less informative about later threshold crossings.
This architecture shift is therefore more difficult for the curve-only predictor than the RL shift observed here.
MARA nevertheless reaches 63.18\%, which is 8.35 points above LARA and 1.11 points above MARA-WLS.
Its larger 7.94-point mean Oracle gap is consistent with the smaller standalone prediction gain and leaves more room for predictors that transfer across model families.

\subsection{Generalization to Heavier Workloads}
\label{sec:load-generalization}

\begin{figure}[t]
  \centering
  \includegraphics[width=\columnwidth]{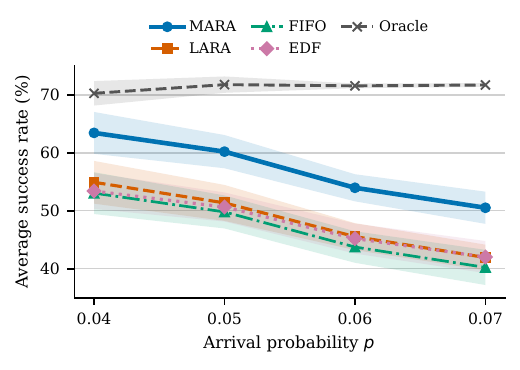}
  \caption{Zero-shot load generalization.
  Lines show means and shaded bands show one sample standard deviation.
  Each run first averages 12 episodes within a task set and then gives equal weight to ID, RL OOD, and ViT OOD.
  MAT is trained only at $p=0.04$.}
  \Description{Line plot of average success rate against arrival probability for MARA, LARA, FIFO, EDF, and the independent-feasibility Oracle.
  MARA is highest among realizable policies at every load, while all realizable policies decline as arrivals become more frequent.}
  \label{fig:load}
\end{figure}

We then evaluate MARA as the task load increases.
Raising the generation probability from 0.04 to 0.07 increases the average number of tasks per episode from 19.70 to 31.75.
Figure~\ref{fig:load} evaluates the fixed $p=0.04$ MAT checkpoints without further tuning at the heavier loads.
MARA declines from 63.46\% at $p=0.04$ to 50.53\% at $p=0.07$, a decrease of 12.93 percentage points despite a roughly 61\% increase in the number of tasks.
Its margin over the strongest main baseline, LARA, remains between 8.43 and 8.84 points across all four settings.
This persistent separation provides evidence of load generalization rather than exposure to a mixed-load training distribution.

Oracle remains near 70--72\% because it evaluates every task independently.
Its mean gap to MARA grows from 6.83 points at $p=0.04$ to 21.19 points at $p=0.07$, but the widening gap does not imply that individual tasks become intrinsically harder.
Instead, more independently feasible tasks compete for the same seven nodes, so admission blocking and queueing account for a larger share of failures.
The nearly flat Oracle curve provides a useful control because task generation leaves independent feasibility largely unchanged while causing more service windows to overlap.
As more tasks approach their deadlines simultaneously, each assigned batch carries a larger opportunity cost.
This is precisely the regime in which remaining-demand forecasts and coordinated node choices should matter, and MARA remains effective as both sources of contention intensify.

\subsection{Operational Analysis}
\label{sec:operational-analysis}

\begin{table}[t]
  \caption{Operational diagnostics over all loads and task sets, reported as mean $\pm$ sample standard deviation.
  Ready is the number of steps from arrival to four loss observations.}
  \label{tab:diagnostics}
  \centering
  \setlength{\tabcolsep}{3.5pt}
  \begin{tabular}{@{}lcccc@{}}
    \toprule
    Method & Util. (\%) & Queue full (\%) & Waiting & Ready \\
    \midrule
    MARA & $73.8\pm2.7$ & $34.8\pm4.7$ & $1.25\pm0.13$ & $44.8\pm10.8$ \\
    MARA-WLS & $73.9\pm2.7$ & $36.2\pm5.2$ & $1.30\pm0.15$ & $46.9\pm12.4$ \\
    FIFO & $74.9\pm2.9$ & $42.1\pm4.8$ & $1.44\pm0.14$ & $88.1\pm11.0$ \\
    EDF & $74.6\pm2.9$ & $41.2\pm4.7$ & $1.42\pm0.14$ & $87.1\pm11.4$ \\
    \bottomrule
  \end{tabular}
\end{table}

To understand why MARA outperforms the baselines beyond aggregate success rates, we measure utilization, queue congestion, waiting population, and the time required to collect enough observations for prediction.
Table~\ref{tab:diagnostics} rules out a simple utilization explanation because all four methods keep nodes busy for roughly 74--75\% of simulator steps.
Relative to FIFO and EDF, MARA reduces the mean full-queue fraction by 7.3 and 6.4 points, lowers the waiting population by 0.19 and 0.17 tasks, and brings tasks to prediction readiness 43.3 and 42.3 steps earlier.
It therefore spends less of an episode with admission blocked while still allocating enough early service to reveal useful loss histories.
MARA also has a 1.4-point lower full-queue fraction and reaches four observations 2.1 steps sooner than MARA-WLS, while completing 1.07 points more tasks when averaged over all four loads.
Predictor-ready time is a scheduling outcome rather than inference latency because both predictors require four observations and their sub-millisecond computation is negligible on the simulator time scale.
The difference therefore reflects which tasks the policy probes and continues.
Together with the load-dependent Oracle gap, the lower congestion and readiness time indicate that MARA benefits from selective use of constrained capacity rather than higher raw utilization.

\subsection{Ablation Study}
\label{sec:ablation}

\paragraph{Predictor and scheduler}
We first separate the effects of the predictor and allocator through controlled component changes.
The ablation columns in Table~\ref{tab:main} keep these comparisons visible without treating them as primary baselines.
Replacing FM with WLS in MAT lowers the average rate from 63.46\% to 62.07\%, and replacing WLS with FM in the fixed LARA allocator raises the rate from 54.93\% to 60.64\%.
These comparisons show that improved forecasts affect downstream allocation rather than only the standalone regression metric.
With the same FM predictor, MARA is 2.83 points above LARA-FM on average and remains higher in all three task sets.
The two component effects are not additive because predictor errors and scheduling decisions interact.

\paragraph{Reward shaping}
We next test whether dense reward shaping improves MAT training relative to a task-level 0/1 reward observed only when a task terminates.
Table~\ref{tab:reward-ablation} shows that dense shaping raises the average success rate from 57.87\% to 63.46\%, with gains of 5.46, 15.72, and 5.61 percentage points on ID, RL OOD, and ViT OOD, respectively.
The 5.59-point average improvement indicates that intermediate progress provides more informative credit for earlier allocation decisions than terminal completion alone.

\begin{table}[t]
  \caption{Reward shaping ablation for MARA.
  Cells report mean $\pm$ sample standard deviation of success rate (\%).}
  \label{tab:reward-ablation}
  \centering
  \begin{tabular}{@{}lcc@{}}
    \toprule
    Task set & MARA & MARA w/ sparse reward \\
    \midrule
    ID & \textbf{61.90 $\pm$ 4.09} & $56.44 \pm 3.62$ \\
    RL OOD & \textbf{65.31 $\pm$ 3.18} & $59.59 \pm 2.87$ \\
    ViT OOD & \textbf{63.18 $\pm$ 3.92} & $57.57 \pm 3.26$ \\
    \midrule
    Average & \textbf{63.46 $\pm$ 3.61} & $57.87 \pm 3.14$ \\
    \bottomrule
  \end{tabular}
\end{table}

\FloatBarrier

\section{Conclusion}
\label{sec:conclusion}

We studied CoRE-Learning with dynamic task arrivals, discrete compute nodes, and unknown training demands.
MARA combines conditional flow matching with an autoregressive multi-agent allocator and uses a log-progress potential to provide dense feedback without changing the completion objective.
At the training load, MARA completes 63.46\% of tasks, versus 54.93\% for LARA, and remains ahead on held-out task families and unseen heavier workloads.
Component ablations confirm that improved resource estimates drive end-to-end gains.

Our simulator assumes homogeneous nodes and cost-free preemption and migration, omitting device heterogeneity and cluster overheads.
Optimizer and data-order effects may further widen the deployment gap.
Although no private data or high-impact decisions are involved, allocation can delay difficult task families, motivating group-level audits and fairness constraints.
Future work will incorporate heterogeneity, switching costs, fairness, and real-cluster validation.

\bibliographystyle{ACM-Reference-Format}
\bibliography{references}

\appendix
\section{Implementation and Experimental Details}
\label{app:implementation}

\subsection{Task Parameters}

This subsection records how the inherited task configurations are adapted to the discrete simulator.
Table~\ref{tab:tasks} lists the resulting calibrated parameters.
For ID arrivals, the listed $\epsilon$ is multiplied by a value from $\mathcal{U}[0.9,1.1]$, and available time is sampled uniformly from the listed integer interval.
RL and ViT sample $\epsilon$ log-uniformly over the listed range and available time uniformly from the listed set.
The task models, datasets, and original task bundles follow LARA~\cite{wang2024lara}, while the target losses and service budgets are recalibrated for cumulative-average, mean-reduced losses and discrete service.

\begin{table*}[t]
  \caption{Learning tasks, calibrated $\epsilon$ values, and effective available-time ranges.
  Available time is measured in simulator steps.
  The raw configuration stores twice the reported value because the simulator applies a fixed $0.5$ time scale.}
  \label{tab:tasks}
  \centering
  \begin{tabular}{@{}llllcc@{}}
    \toprule
    Split & Task configuration & Model & Dataset & $\epsilon$ & Available time \\
    \midrule
    ID & \texttt{CNN\_Cifar10} & CNN classifier & CIFAR-10 & $1.6601$ & $[285,315]$ \\
    ID & \texttt{ResNet18\_Cifar10} & ResNet-18 & CIFAR-10 & $1.8575$ & $[285,315]$ \\
    ID & \texttt{ResNet34\_Cifar10} & ResNet-34 & CIFAR-10 & $1.8523$ & $[285,315]$ \\
    ID & \texttt{LSTM\_Cifar10} & LSTM classifier & CIFAR-10 & $2.1271$ & $[285,315]$ \\
    ID & \texttt{TSFM\_Audio\_1} & Transformer encoder & YESNO & $0.4762$ & $[292,350]$ \\
    ID & \texttt{TSFM\_Audio\_2} & Transformer encoder & YESNO & $0.6043$ & $[292,350]$ \\
    ID & \texttt{ALSTM\_Text\_1} & Attention LSTM & IMDB & $0.0249$ & $[312,345]$ \\
    ID & \texttt{ALSTM\_Text\_2} & Attention LSTM & IMDB & $0.0214$ & $[312,345]$ \\
    OOD & \texttt{RLExpertConfig} & DAggerCNNNet & Montezuma's Revenge & $[1.4,1.8]$ & $\{285,295,305,315\}$ \\
    OOD & \texttt{ViT\_Cifar10\_1} & SimpleViT & CIFAR-10 & $[2.0231,2.0440]$ & $\{285,295,305,315\}$ \\
    \bottomrule
  \end{tabular}
\end{table*}

\subsection{Predictor Settings}

The predictor settings cover both curve generation and flow-model inference.
The FM dataset contains 480 complete ID curves, with 50 training curves and 10 validation curves for each of the eight configurations.
Each task runs to twice its scheduling budget.
Image, audio, and text batch sizes are sampled from $\{16,32,64\}$, $\{1,2,4\}$, and $\{4,8\}$, respectively.
The base learning rate is log-uniform on $[3\!\times\!10^{-4},3\!\times\!10^{-3}]$.
All tasks use Adam, zero weight decay, no learning-rate scheduler, fixed-size complete mini-batches, and mean-reduced losses.

The predictor uses a 64-dimensional GRU history encoder and a two-layer MLP velocity field.
ID and RL evaluation uses eight flow samples and four Euler steps, whereas ViT uses four samples and one Euler step selected on predictor validation data.
WLS uses forgetting factor $0.9$.

\subsection{Scheduler and Simulator Settings}

The remaining settings specify scheduler training and simulator execution.
MAT is trained for 100,000 environment steps with three training seeds and arrival probability $p=0.04$.
Each run uses one Transformer block, 64-dimensional embeddings, and one attention head.
PPO uses clipping coefficient $0.2$, value coefficient $c_v=1$, entropy coefficient $c_e=0.01$, discount $\gamma=1$, and GAE parameter $\lambda=0.95$.
The shaping coefficient is $\beta=0.1$.
Checkpoint step 10,500 was fixed before evaluation.

The simulator contains seven homogeneous nodes, admits at most ten active tasks, and runs for 500 steps.
One selected task executes one optimization batch.
Evaluation uses five independently generated roots.
At $p=0.04$, the same 12 pre-generated episodes and task curves are used by every method.
Higher-load evaluation keeps the trained checkpoint fixed and changes only $p$ to $0.05$, $0.06$, or $0.07$.

\end{document}